\documentclass{amia}
\usepackage[dvipsnames]{xcolor}

\usepackage{graphicx}
\usepackage[labelfont=bf]{caption}
\usepackage[superscript,nomove]{cite}
\usepackage{color}

\usepackage{latexsym}
\usepackage{amsmath}
\usepackage[colorinlistoftodos]{todonotes}
\usepackage{url}
\usepackage{multicol}

\begin{document}

\title{Towards more patient friendly clinical notes\\through language models and ontologies}

\author{
Francesco Moramarco, Damir Juric, Aleksandar Savkov, Jack Flann, Maria Lehl, Kristian Boda, Tessa Grafen, Vitalii Zhelezniak, Sunir Gohil, Alex Papadopoulos Korfiatis, Nils Hammerla
}

\institutes{
    Babylon Health, London, UK\\
}

\maketitle

\noindent{\bf Abstract}

\textit{
  Clinical notes are an efficient way to record patient information but are notoriously hard to decipher for non-experts. 
  Automatically simplifying medical text can empower patients with valuable information about their health, while saving clinicians time.
  We present a novel approach to automated simplification of medical text based on word frequencies and language modelling, grounded on medical ontologies enriched with layman terms.
  We release a new dataset of pairs of publicly available medical sentences and a version of them simplified by clinicians. Also, we define a novel text simplification metric and evaluation framework, which we use to conduct a large-scale human evaluation of our method against the state of the art.
  Our method based on a language model trained on medical forum data generates simpler sentences while preserving both grammar and the original meaning, surpassing the current state of the art.
}
  


\section*{Introduction}
\label{sec:intro}

Making medical information available for patients is becoming an important aspect of modern healthcare, but the frequent use of medical terminology makes it less accessible for patients/consumers. There is a trade-off between promoting more ``patient-friendly'' medical notes \cite{amrc} and the efficiency of clinicians who often prefer writing in shorthand. This is an opportunity for automation, as Natural Language Processing (NLP) and Natural Language Generation (NLG) techniques have the potential to simplify medical text and thereby increase the accessibility to patients while maintaining efficiency. 

Text simplification in the general domain has improved greatly with the introduction of new deep-learning methods borrowed from the field of Machine Translation \cite{Stajner2018}. However, the challenges in medical text simplification are particularly focused around explaining the abundant terminology, much of which is in Greek or Latin \cite{keselman2012classification}. This is why most efforts in the field are concentrated around the use of a mapping table from complex to simple terms \cite{van2019evaluating,shardlow2019neural}.
While the task of language simplification is not new, there are very few datasets specifically built for it \cite{Xu2015}. In the case of medical text simplification, the community has not yet been able to use a common benchmark due to data access constraints \cite{shardlow2019neural}. Perhaps, the only resource that comes close is a medically themed subset of Simple Wikipedia \cite{van2019evaluating,Kauchak2013}. In the context of clinical notes, medical accuracy and safety are of utmost importance, which makes consistent evaluation a strong requirement for sustainable improvements in the field.


We present a medical text simplification benchmark dataset of $1{\,}250$ parallel complex-simple sentence pairs based on publicly available medical sample reports. Furthermore, we propose a novel approach to lexical simplification for the medical domain, which uses a comprehensive ontology of medical terms and their alternatives, and a novel scoring function that combines language model (LM) probabilities and word frequencies into one unified measure.
We conduct a human evaluation to validate our method and find that unbounded, left-to-right LMs trained on medical forum data achieve the best results on our benchmark dataset. Finally, we make the source code for our method, and all materials necessary to repeat the human evaluation, available on GitHub\footnote{\url{https://github.com/babylonhealth/laymaker}}. While evaluated in the medical domain, this approach can be abstracted into other domains by utilising an appropriate alternative ontology and suitable language model training data.

Our contributions are the following: a dataset of simplified medical sentences, a new approach for text simplification, an evaluation framework for text simplification, and a model that generates simpler, grammatically correct sentences with their original meaning preserved.


\section*{Related work} 
\label{sec:related_work}

\paragraph{General text simplification.} Initial efforts on automatic text simplification use Phrase-based Machine Translation (PB-MT) methods \cite{10.1007/978-3-642-12320-7_5} driven by the availability of two resources: the open-source framework Moses \cite{moses} and the Simple English Wikipedia dataset \cite{coster_data}. These early PB-MT systems perform well, but remain too careful in suggesting simplifications. Later work provides extensions that address some of these issues --- deletion  \cite{coster_lm} and Levenshtein distance based ranking \cite{wubben}. Stajner et al. (2015)\cite{vstajner2015deeper} provide an insight into how much of an effect the size and the quality of the training data has on the performance of the MT systems.

Machine translation algorithms trained on parallel monolingual corpora, such as the Newsella\footnote{\url{https://newsela.com/data}} parallel corpus,  have shown great promise in recent years \cite{wang2016text}, combining, ideally, lexical and syntactic simplification. Nisioi et al. (2017)\cite{nisioi-etal-2017-exploring} use the OpenNMT package \cite{opennmt} to simultaneously perform lexical simplification and content reduction. Sulem et al. (2018b)\cite{sulem2018simple} show that performing sentence splitting based on automatic semantic parsing in conjunction with neural text simplification (NTS) improves both lexical and structural simplification.



\paragraph{Medical text simplification.} A complex vocabulary is typically the main hindrance to understanding medical text, and is therefore the main target for simplification \cite{shardlow2014survey}. Fortunately, there are numerous medical ontologies containing multiple ways of expressing the same medical term, often including an informal, layman alternative \cite{nelson2017unified, kohler2018expansion}. Using these ontologies to replace complicated words with more common ones is a recurring theme in medical text simplification \cite{abrahamsson2014medical,van2019evaluating,shardlow2019neural}.
Abrahamsson et al. (2014)\cite{abrahamsson2014medical} show a preliminary study on a method that replaces specialised words derived from Latin and Greek with compounds from every-day Swedish words, and achieve encouraging results on readability. Shardlow et al. (2019)\cite{shardlow2019neural} use existing neural text simplification software augmented with a mapping between complex medical terminology and simpler vocabulary taken from the alternative text labels of SNOMED-CT. Their simplification method has an increased understanding among human evaluators based on a crowd-sourced evaluation process. Van den Bercken et al. (2019)\cite{van2019evaluating} use a neural machine translation approach that is aided by a terminology--mapping table that decreases the medical vocabulary in the (complex) source text.

Despite these efforts, the field still lacks a benchmark dataset based on real medical data as well as accessible open source medical baselines; the exception being the small, medically themed subset of Simple Wikipedia provided by Van den Bercken et al. (2019)\cite{van2019evaluating}. The main drawback of this corpus is that it tends to simplify sentences by omitting some of the information, which is not a viable method in the context of clinical notes. Medical data is highly sensitive and even its use for research purposes is strictly regulated and often difficult. Therefore a new medical data resource is bound to have a great impact and move the field forward, as it has happened in the past \cite{Uzuner2011,Johnson2016}.
\section*{Dataset}
\label{sec:data}

The \emph{MTSamples} dataset comprises around $5{\,}000$ sample medical transcription reports from a wide variety of specialities uploaded to a community platform website\footnote{\url{https://mtsamples.com}}. However, publicly available annotations are limited to only include high-level metadata, e.g. the medical speciality of a report.

We create a parallel corpus of clinician-simplified medical sentences on the basis of the raw \emph{MTSamples} dataset. We pre-process the entire original dataset by tokenising all sentences and expanding abbreviations based on a custom list of common medical ontologies compiled by clinicians. We then review and exclude sentences that have too little context (i.e. are confusing or ambiguous to a clinician) or grammatically incorrect. Finally, three clinicians (native British English speakers) create a new version of each sentence using layman terms, ensuring consistency of both structure and medical context and accuracy. Only one simple sentence is generated for each original sentence for which simplification is possible. The resulting dataset contains $1{\,}250$ sentence pairs, of which $597$ ($47.76\%$) have been simplified. The remainder have been left unchanged because they could not be further simplified. The average number of tokens in the original sentences is $66.96$, and in the simplified sentences $68.60$.

We divide the data into a $250$ sentence development set and a $1{\,}000$ sentence test set. 

\section*{Medical ontologies}
\label{sec:ontologies}

Recognising concepts and subsequently linking with a medical ontology are common medical NLP tasks necessary for higher level analysis of medical data \cite{zheng2015entity}. Many semantic tagging systems use the labels (text representations) defined as part of the ontologies to recognise possible instances of the entities in the text. Typically, every concept has a primary official label as well as at least a few alternative labels. Ideally these labels should be interchangeable; thus, they can be used to replace more complicated labels with layman alternatives.

In order to maintain good coverage of both medical terms and layman terminology, we select three state-of-the-art medical ontologies for creation of our phrase table. \emph{SNOMED-CT} is one of the most comprehensive medical terminologies in the world, and is also available in different languages. As of the January 2019 release, it comprises $349{\,}548$ medical concepts, covering virtually all medical terminology used by clinicians. We also include the Consumer Health Vocabulary (CHV), the purpose of which is lexical simplification \cite{zeng2006exploring}, and the Human Phenotype Ontology (HPO), which is a standardized vocabulary of phenotypic abnormalities encountered in human disease, and also contains a layer of plain language synonyms \cite{PMID:29632381}.

We create a \textit{vocabulary} of medical terms (named entities) based on the labels of concepts from these ontologies --- approximately $460{\,}000$ labels from $160{\,}000$ concepts. For example, the concept label ``Otalgia'' has alternative labels ``Pain in ear'' (Snomed), ``Earache'' (CHV), and ``Ear pain'' (HPO). To produce it, we align the ontologies using the union-find algorithm \cite{patwary2010experiments} and discard duplicate labels, as well as those without alternatives, as they cannot contribute to the simplification process.

\section*{Lexical simplification}
\label{sec:approach}

\begin{figure}[t]
    \centering
    \includegraphics[width=1.0\textwidth]{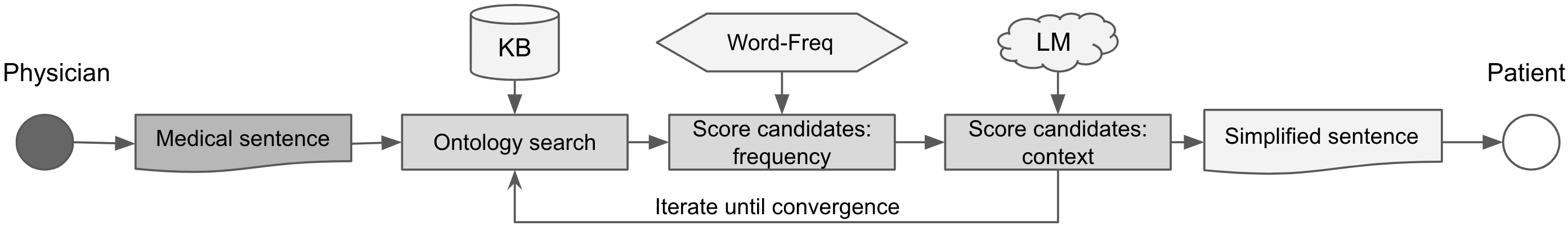}
    \caption{A flow diagram of the simplification algorithm.}
    \label{fig:flow-diagram}
\end{figure}

Lexical text simplification looks to identify difficult words and phrases and replace them with alternatives based on some measure of simplicity. Word frequency over a large amount of text is often chosen as this measure and has been used to both identify and replace candidates \cite{van2019evaluating}. The probability score of a sentence based on some language model has also been used to rank candidates \cite{xu2016optimizing}. Additionally, in the medical domain, terminology words are often assumed to be the main target of lexical simplification \cite{van2019evaluating,shardlow2019neural}.
We propose a new approach to medical lexical text simplification, which uses a vocabulary based on a medical ontology to identify candidates. It then ranks each alternative using a linear combination of word frequency and the sentence score produced by a language model. After completing the replacement and ranking steps for each medical term (of one or more words) in an input sentence, the process is repeated until no further changes are suggested. Figure \ref{fig:flow-diagram} shows a high-level view of the algorithm.

\subsection*{Candidate ranking}
\label{sec:ranking}

The main task of lexical text simplification is to make the overall sentence simpler, so a ranking function should aim to provide the simplest replacement for each entity. However, this introduces a second challenge -- maintaining correct grammar after the replacement. A good ranking function should therefore optimise for both the simplicity and grammaticality of the result.

\vspace{-.1in}

\paragraph{Word frequency} is a strong indicator for simplicity \cite{paetzold-specia-2016-semeval} as it directly measures how common a given word is. However, there are different approaches to how it is utilised for multi-word expressions. Common approaches include taking the \emph{average} \cite{shardlow2019neural}, the \emph{median}, or the \emph{minimum} word frequency. We choose the \emph{minimum}, under the assumption that the least frequent word in the sequence drives the overall understanding of the sequence. For example, consider the candidates \emph{otalgia of ear}, and \emph{earache}. An average or median frequency would score option 1 as simpler because of the very common word \emph{of}, whereas the minimum word frequency would score option 2 as more frequent.
To calculate the word frequencies ($\mathit{WF}$) for a given set of candidate labels, we use the \emph{wordfreq} python package\footnote{\url{http://pypi.org/project/wordfreq/2.2.1}} which provides word frequency distributions calculated over a large general purpose corpus.
\begin{equation*}
  P(w) = \frac{C(w)}{|W|}
\end{equation*}


where $C(w)$ is the number of times the word occurs in the corpus and $|W|$ is the number of words in the corpus. Given a sequence, such as \emph{heart attack}, composed of words $w_1{\dots}w_k$ we calculate its word frequency score, $\mathit{WF}$, as

\vspace{-.1in}

\begin{equation*}
  \mathit{WF}(w_1{\dots}w_k) = min_{i=1}^k \ln{(P(w_i) + \epsilon)}
\end{equation*}
As we seek to combine this probability with language model scores, it makes sense to convert it to a logarithmic scale to avoid computational underflow. For the same reason, we introduce Laplace smoothing \cite{chen1999empirical} through the addition of the constant $\epsilon$ ($10^{-10}$).

\vspace{-.1in}

\paragraph{Language models} have made impressive strides in recent years, showing that they are capable of generating complex syntactic constructions while maintaining good grammar and coreference \cite{gulordava2018colorless,radford2019language}. We argue that the latter quality makes them a good predictor of grammatical correctness. Given that lexical simplification relies on the replacement of a recognised span from the sentence with a simpler one from a vocabulary, language models can be used to determine a score for how well a new term conforms to the grammar of the sentence.

To calculate this score we train a language model on a dataset of $160{\,}000$ original, top-level posts (1.8M sentences), scraped from the Reddit's AskDocs\footnote{\url{https://www.reddit.com/r/AskDocs/}} forum. This dataset contains sentences which are largely medical and therefore will have the necessary vocabulary, while its language style is predominantly layman since the top post in a thread is usually written by a non-expert looking for medically-related information.

Given a sequence of words $w_1{\dots}w_n$ and a language model, we can estimate the likelihood of the sequence as the log-probability of each word occurring given all preceding words in the sentence:
\begin{equation*}
  \ln{\hat{P}(w_1{\dots}w_n)} = \frac{1}{n} \sum_{i=1}^n \ln{P(w_i|w_{i-1},{\dots},w_1, \langle s \rangle)}
\end{equation*}
where $\langle s \rangle$ is the start symbol and $n$ the number of tokens in the sequence. We normalise by the number of tokens $n$ to account for replacement terms of different length, e.g. \emph{dyspnoea} and \emph{shortness of breath}.
The language model gives a signal for how appropriate and grammatically correct the replacement term in the given sentence is.
Table \ref{tab:sentence_scores} shows both the language model $\mathit{LM}$ and frequency $\mathit{WF}$ scores for the term replacements of \emph{myocardial infarctions}.
In our example, the $\mathit{LM}$ scores \emph{heart attacks} (notice the plural) above \emph{heart attack} given the context \emph{Patient had multiple}.
Given the frequency score ($\mathit{WF}$) of a replacement term ($T_i$) and the language model score ($\mathit{LM}$) for its corresponding replacement sentence ($S'_i$), we define the final score as a linear combination of the two:
\begin{equation}
  \text{Score}(T_i) = \alpha \mathit{LM}(S'_i) + (1 - \alpha) \mathit{WF}(T_i)
  \label{equation:final_score}
\end{equation}
We then select the term with the highest score. The parameter $\alpha \in [0,1]$ acts as a regulariser and can be fine-tuned on a separate dataset. When $\alpha = 0$, the score is entirely driven by $\mathit{WF}$. When $\alpha = 1$, the score is entirely driven by $\mathit{LM}$. We select suitable $\alpha$ values on the development set.

\begin{table}[t]
    \small
    \begin{minipage}{0.5\linewidth}
    \caption{LM and Frequency scores for alternative labels\\of \emph{myocardial infarctions.}}
      \centering
        \begin{tabular}{ |l||c|c| }
            \hline
            \textbf{Candidate} & \textbf{LM} & \textbf{Freq.} \\
            Patient had multiple... & \textbf{Score} & \textbf{Score} \\
            \hline
            \emph{myocardial infarctions} & -5.45 & -14.32 \\
            \emph{heart attack} & -4.38 & \textbf{-9.05} \\
            \emph{heart attacks} & \textbf{-3.91} & \textbf{-9.05} \\
            \emph{mies} & -6.09 & -14.34 \\
            \emph{myocardial necrosis} & -6.13 & -14.23 \\
            \hline
        \end{tabular}
        \label{tab:sentence_scores}

    \end{minipage}%
    \begin{minipage}{.5\linewidth}
      \centering
    \caption{An example of the language model (LM) convergence.}
    \small
    \begin{tabular}{|p{1.3cm}|p{5.8cm}|}
         \hline
        \textbf{Iteration} & \textbf{Sentence} \\
        \hline
        Original    & hyperlipidemia with elevated triglycerides . \\
        \hline
        Iteration 1 & \textcolor{RoyalBlue}{elevated lipids in blood} \textcolor{RoyalBlue}{in addition to} \textcolor{RoyalBlue}{high} triglycerides .\\
        \hline
        Iteration 2 & \textcolor{RoyalBlue}{excessive fat} in \textcolor{RoyalBlue}{the} blood \textcolor{RoyalBlue}{with} high triglycerides .\\
        \hline
    \end{tabular}
        \label{tab:convergence}

    \end{minipage} 
\end{table}

\subsection*{Simplification algorithm}
\label{sec:algorithm}
A comprehensive vocabulary often results in overlapping candidate spans. For example, in the sentence \emph{Patient has lower abdominal pain}, the following 5 spans match an entity: \emph{lower}, \emph{abdominal pain}, \emph{abdominal}, \emph{pain}, and \emph{lower abdominal pain}.
In the case where two or more spans overlap or one is subsumed by the other, the algorithm takes a greedy left-to-right processing approach. It ranks the spans in order from left to right, prioritising longer spans and ignoring all spans that have any overlap with an already processed span. Additionally, it is fairly common for a sentence to contain more than one non-overlapping medical terms. For example, consider the artificial sentence: \emph{Patient has a history of myocardial infarction, tinnitus, otalgia, dyspnoea and respiratory tract infection.}, which has multiple, non-overlapping spans suitable for replacement. When constructing candidate sentences to score, replacing only one complex term while leaving the rest of the sentence unchanged yields a sub-optimal score. The optimal approach would be to perform an exhaustive search of all possible combinations within the sentence. Given $n$ terms, and $r$ replacements per term on average, exhaustive search would require $r^n$ combinations, i.e. exponential in the number of terms in the sentence. Rather than introduce this computational cost, we instead consider each term independently of the others. After simplifying all of them, we repeat the extract-and-replace process (see Figure \ref{fig:flow-diagram}) until no further change occurs, i.e. until convergence (see Table \ref{tab:convergence}). 
This reduces the time complexity to $n \cdot t$, where $t$ is the number of iterations to reach convergence. We cap the number of iterations $t$ at 5, as our experiments show only 1 out of $1{\,}000$ sentences to ever reach this many iterations. In practice, we find that most sentences converge after one iteration, with a median of $1$ iterations and an average of $1.19$.
\section*{Experimental setup}
\label{sec:setup}

Our method requires a language model to score alternative terms. To assess the best model for this purpose we train three different language models. Next, we fine-tune $\alpha$ for each of them and proceed to measure their respective success against the human-generated reference.
The language models we select are:
\vspace{-.1in}
\begin{itemize}
    \setlength{\itemsep}{0pt}%

    \item \textbf{ngram} — a trigram language model built with KenLM\footnote{\url{https://github.com/kpu/kenlm}} and trained on Reddit AskDocs;
    
    \item \textbf{GPT-1} — a neural language model \cite{radford2018improving} trained on Reddit AskDocs;
    
    \item \textbf{GPT-2} — a neural language model \cite{radford2019language} pretrained only on generic English text. We don't fine-tune this model to evaluate whether general-purpose language models are better at choosing layman alternatives. 
\end{itemize}
\vspace{-.1in}
In order to evaluate our approach, we compare it against three methods from the literature:
\vspace{-.1in}
\begin{itemize}
    \setlength{\itemsep}{0pt}%

    \item \textbf{NTS} — Nisioi et al. (2017) \cite{nisioi-etal-2017-exploring} train an encoder-decoder on Simple Wikipedia, which contains a proportion of medical sentences;
    
    \item \textbf{ClinicalNTS} —  Shardlow et al. (2019)\cite{shardlow2019neural} augment the system by Nisioi et al. (2017) \cite{nisioi-etal-2017-exploring} with a medical phrase table, which is the current state of the art for clinical text simplification;
    
    \item \textbf{PhraseTable} — a simple term replacement system based on the phrase table from Shardlow et al. (2019) \cite{shardlow2019neural}, which we consider our baseline.
\end{itemize}
\vspace{-.1in}

\begin{figure}[t]
    \centering
    \includegraphics[width=0.45\textwidth]{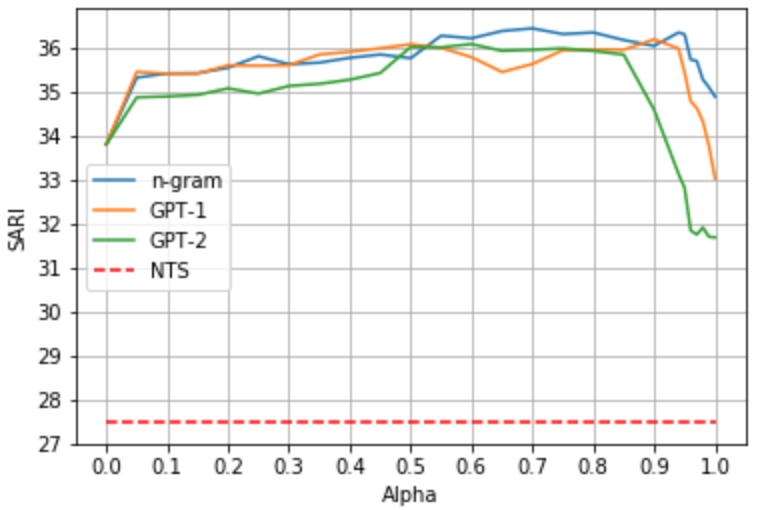}
    \caption{Grid search results for $\alpha$ values between 0 and 1 with a step of 0.05. Additional tests with step 0.01 were conducted for values between 0.90 and 1. The best performing $\alpha$ values for each model are 0.70 for the ngram, 0.90 for GPT-1, and 0.60 for GPT-2.}
    \label{fig:alpha}
\end{figure}

The $\alpha$ parameter introduced in Equation \ref{equation:final_score} regulates the ratio of the language model and the word frequency score used for scoring a replacement term. A held-out development set of 250 sentences is used for tuning the $\alpha$ parameter for each of our models. For this purpose we use the automatic metric SARI \cite{xu2016optimizing}, as it intrinsically measures simplicity by comparing the model output against both the human reference and the input sentence.
We perform grid search on the $\alpha$ space ($0$ to $1$) for each model (see Figure \ref{fig:alpha}) and select the top $\alpha$ to be used in the final evaluation.

\subsection*{Traditional evaluation metrics}
\label{sec:autometrics}

There are three general evaluation approaches for simplification that have been tried in the past:
\vspace{-.1in}
\begin{itemize}
    \setlength{\itemsep}{0pt}%
    \item BLEU score \cite{Papineni:2002:BMA:1073083.1073135} is one of the standard metrics of success in machine translation and has been used in some cases for simplification \cite{zhu2010monolingual} as it correlates with human judgements of meaning preservation.
    \item SARI is a lexical simplicity metric that measures the appropriateness of words that are added, deleted, and kept by a simplification model \cite{van2019evaluating,nisioi-etal-2017-exploring}.
    \item Human evaluation, either through dedicated annotators or crowd-sourcing, indicating whether the generated sentences are considered simpler by the end users. 
\end{itemize}
\vspace{-.1in}

\noindent
Both SARI and BLEU are intended to have multiple references for each sentence to account for syntactic differences in the simplified text. As we only have one simplified reference for each original sentence, these metrics are likely to be somewhat biased to a particular way of expression. Therefore, conducting a human annotation process can bring additional reassurance to the evaluation process.

\subsection*{Human annotation}
\label{sec:annotation}

We design a human evaluation process in the form of a crowd-sourced annotation task on Amazon Mechanical Turk (MTurk) \cite{doi:10.1177/1745691610393980}. The goal of the task is to determine whether a simplified sentence is better than the original.
 Celikyilmaz et al. (2020) \cite{celikyilmaz2020evaluation} identify the two most common ways to conduct human evaluation on generated text: (i) ask the annotators to score each simplified sentence independently with a Likert scale, (ii) ask the annotators to compare sentences simplified by different models. We experiment with both methods and decide to opt for the latter, which produces more consistent results, as also shown by Amidei et al. (2019) \cite{amidei2019use}. For this purpose, we create sentence pairs from each original sentence (marked as A) and either a sentence simplified by the model or the gold simplification provided in the dataset (marked as B). We use the following four categories:

\begin{tabular}{ll}
    \centering
    1. Sentence A is easier to understand. & 2. Sentence B is easier to understand. \\
    3. I understand them both the same amount. & 4. I do not understand either of these sentences.
\end{tabular}


\noindent
Often, the simplified sentence generated by the models is identical to the original sentence. To save annotation resources we annotate such pairs only once and extrapolate the annotation to all models. MTurk provides little control over the reading age and language capabilities of the annotators, so we have to account for some variability in the annotation. Therefore, all sentence pairs are annotated 7 times by different annotators. In total, the annotations comprise $20{\,}965$ sentence pairs derived from $2{\,}995$ unique ones. Finally, we use the option of selecting only ``master'' annotators\footnote{Master annotators are annotators whose work has not been rejected by task requesters for some period of time.} for the task, as it is difficult to judge the quality of the work of particular annotators. We choose turkers without medical experience, as opposed to medical professionals, because they are a good representation of the end users of such system. We assume that the human reference should both succeed more often and fail less often than any of the models. We measure the quality of the models with a Simplification Gain $\mathit{SG}$ that we define as the difference between successes $S$ (option 2.) and the failures $F$ (option 1.), normalised by the total number of pairs, $T$:

\begin{table}[t]
    \centering
    \caption{Human judgement counts for sentence pairs from the test set for all models and the reference human sentences. \textbf{S:} the generated was simpler; \textbf{F:} the original was simpler; \textbf{E:} both of equal complexity; \textbf{N:} cannot understand either; \textbf{U:} was not changed by the model/human reference; \textbf{SG} simplification gain as defined in Equation \ref{eq:ratio}. Bold indicates best model. Scores in \textbf{SG} are significant ($p<0.05$)}
    \setlength\tabcolsep{4.2pt} 
    \begin{tabular}{c|ccccc|c}
              \multicolumn{1}{c}{~} &
              \multicolumn{1}{c}{\textbf{S}}    & \multicolumn{1}{c}{\textbf{F}}    & \multicolumn{1}{c}{\textbf{E}}    & \multicolumn{1}{c}{\textbf{N}}   & \multicolumn{1}{c}{\textbf{U}}    & \multicolumn{1}{c}{\textbf{SG}}    \\
        \hline
        \hline
        Human & \textcolor{OliveGreen}{$1{\,}730$} & \textcolor{BrickRed}{273}  & 904  & 40  & {$4{\,}053$} & \textcolor{NavyBlue}{0.21} \\ \hline
        n-gram     & \textcolor{OliveGreen}{${1{\,}452}$} & \textcolor{BrickRed}{$1{\,}004$} & {$1{\,}732$} & 110 & {$2{\,}702$} & \textcolor{NavyBlue}{0.06}  \\
        GPT-1     & \textcolor{OliveGreen}{$1{\,}404$} & \textcolor{BrickRed}{\textbf{747}}  & {$1{\,}736$} & 117 & {$2{\,}996$} & \textcolor{NavyBlue}{\textbf{0.09}} \\
        GPT-2     & \textcolor{OliveGreen}{$1{\,}372$} & \textcolor{BrickRed}{$1{\,}077$} & {$1{\,}661$} & 118 & {$2{\,}772$} & \textcolor{NavyBlue}{0.04}\\ \hline
        NTS       & \textcolor{OliveGreen}{587}  & \textcolor{BrickRed}{855}  & $1{\,}022$ & 98  & {$4{\,}438$} & \textcolor{NavyBlue}{-0.04} \\
        ClinicalNTS       & \textcolor{OliveGreen}{1{\,}483}  & \textcolor{BrickRed}{1{\,}597}  & 404 & \textbf{93}  & {$3{\,}423$} & \textcolor{NavyBlue}{-0.02} \\
        PhraseTable       & \textcolor{OliveGreen}{$\mathbf{2{\,}425}$}  & \textcolor{BrickRed}{2{\,}759}  & $\mathbf{269}$ & 98  & {$1{\,}449$} & \textcolor{NavyBlue}{-0.05} \\ \hline
    \end{tabular}
    \label{tab:human_eval}
\end{table}

\begin{equation}
    \label{eq:ratio}
   \mathit{SG} = \frac{S-F}{T}
\end{equation}

\section*{Results}
\label{sec:evaluation}

We count all judgements of the same category for each model and the human reference, and present the results in Table \ref{tab:human_eval}. Additionally, based on these counts we calculate the simplification gain $\mathit{SG}$ as described in Equation \ref{eq:ratio}. We can make the following conclusions based on this data:

\vspace{-.1in}

\begin{itemize}
    \setlength{\itemsep}{0pt}%
    
    \item the human reference is very rarely more complex than the original, which makes a considerable difference in its Simplification Gain, as opposed to most of the models, which seem to be prone to this kind of error (see columns \textbf{F} and \textbf{SG});
    
    \item based on the Simplification Gain in \textbf{SG}, the GPT-1 model yields the best performance.
    We believe this is due to: (i) having access to the entire context (as opposed to n-gram), which makes it cautious about simplification, and (ii) being more focused on medical terminology due to its training set (as opposed to GPT-2);
    
    \item the methods we compare against have a negative Simplification Gain, meaning the number of failures exceeds the number of successes. General-purpose NTS is less eager in its simplification (column U in Table \ref{tab:human_eval}), which could be explained by the divergence between its training set (Simple Wikipedia) and our test set (Clinical Notes). 
    Both ClinicalNTS and PhraseTable overcome this by applying a medical phrase table (see Section 6 for more details), which triggers more medical replacements. ClinicalNTS has higher Simplification Gain overall compared to general purpose NTS, which is to be expected, but still fails more often than succeed;
    
    \item A possible explanation for the high number of successes of NTS and ClinicalNTS lies in their aggressive removal of phrases, which makes them easier to understand, but at a considerable loss of information. Both systems use a model trained on Simple Wikipedia, which very often simplifies sentences by removing words or phrases. For example, the original sentence: ``\emph{It has normal uric acid, sedimentation rate of 2, rheumatoid factor of 6, and negative antinuclear antibody and C-reactive protein that is 7.}" is simplified into ``\emph{It has normal uric acid.}"
    
\end{itemize}
\vspace{-.1in}

\noindent
We also report the scores for the most commonly used automatic metrics in the field, BLEU and SARI, though we stress that these scores are unreliable due to (i) their limitations as shown by Sulem et al. (2018) \cite{sulem-etal-2018-bleu} — they only use surface level syntactic features, and (ii) they perform better with multiple references and we only have one.
The NTS baseline is still performing poorly in most metrics except for BLEU, which is likely due to its conservative approach resulting in a large number of unchanged sentences that likely overlap with the reference sentences.


\begin{table}[t]
    \begin{minipage}{.5\linewidth}
        \caption{Reference-based metrics. \textbf{BLEU} and \textbf{SARI}\\calculated using the human-generated reference\\sentences.}
      \centering
        \begin{tabular}{c|ccc}
              \multicolumn{1}{c}{~} &
              \multicolumn{1}{c}{\textbf{BLEU}}    & \multicolumn{1}{c}{\textbf{SARI}}    \\
        \hline
        \hline
        n-gram  & 66.31 & 33.40 &  \\ 
        GPT-1   & 68.19 & \textbf{33.57}  \\ 
        GPT-2   & 66.45 & 33.40 & \\\hline 
        NTS     & \textbf{70.17}  & 27.67 \\ 
        ClinicalNTS     & 68.22 & 30.14   \\
        PhraseTable     & 53.37 & 27.70   \\\hline
    \end{tabular}
        \label{tab:ref_metrics}

    \end{minipage}%
    \begin{minipage}{.5\linewidth}
      \centering
        \caption{Grammaticality and meaning preservation scores over a sample of 1250 generated sentences. \textbf{G1:} no errors; \textbf{G2:} minor errors; \textbf{G3:} major errors; \textbf{M:} the meaning is preserved. Bold implies best result.}
    \begin{tabular}{c|ccc|c}
    \multicolumn{1}{c}{~} & \multicolumn{1}{c}{\textbf{G1}} & \multicolumn{1}{c}{\textbf{G2}}& \multicolumn{1}{c}{\textbf{G3}} & \multicolumn{1}{c}{\textbf{M}} \\
    \hline
    \hline
     \small n-gram & 65.2\% & 25.6\% & 9.2\% & 93.3\% \\
     \small GPT-1 & \textbf{75.4\%} & 17.8\% & \textbf{6.8\%} & \textbf{93.4\%} \\
     \small GPT2 & \textit{69.3\%} & 21.6\% & 9.1\% & 89.6\% \\
     \hline
     \small NTS & \textbf{75.4\%} & \textbf{9.6\%} & 15\% & 63.1\% \\
    \small ClinicalNTS & 43.4\% & 42\% & 14.4\% & 60.4\% \\
    \small PhraseTable & 31.6\% & 34.8\% & 33.6\% & 60.8\% \\
    \hline
    \end{tabular}
    \label{tab:gram}
    \end{minipage} 
\end{table}

\vspace{-.1in}
\paragraph{To test the impact of convergence,} we perform an ablation study on all our models. We take all the sentences that require more than one iteration to converge (around 10\% of the dataset) and perform the same human annotation through Mechanical Turk. Our results show that convergence improves $\textit{SG}$ for all models except GPT-2 by reducing the number of miss-simplified sentences. Empirically we find that the GPT-2 tends to increase the length of the sentence at each iteration, falling into a loop typical for language models.

\subsection*{Grammaticality and meaning preservation}
\label{sec:grammaticality}

Asking end users to rank two sentences in order of simplicity is not enough to judge whether a generative model is performing well. A model should be penalised if the simplified sentence is grammatically incorrect or if it has altered the meaning of the original sentence.
To test these two criteria, we take a random sample of 1250 simplified sentences from all models from the test set. We ask a linguist to assign one of three grammaticality categories: no errors (G1), minor errors (G2), and major errors (G3). We then ask a clinician to mark sentences from the same sample where the meaning has changed in any way.

Table \ref{tab:gram} summarises our findings. It clearly shows the contributions of a good language model in both preserving grammar and meaning. Our method, which is informed by language models, scores highest in both criteria. NTS, which uses a language model decoder, is quite successful in preserving grammar but less successful in preserving meaning. This is likely due to its training set, which encourages the model to remove complex phrases to simplify a sentence. ClinicalNTS and PhraseTable, which rely on a hard-coded phrase table of medical substitutions, score lower both in grammaticality and meaning preservation.

\section*{Conclusion}
\label{sec:conclusion}


In this paper, we present a novel approach to medical text simplification in a effort to empower patients with valuable information about their own health.

First, we address the lack of high quality, medically accurate, and publicly available datasets for evaluating medical text simplification by creating such a dataset with the help of medical professionals. 
Second, we propose an evaluation framework for assessing the quality of simplification algorithms in the medical domain, including an experimental setup for crowd-sourced human evaluation and a metric, which we call Simplification Gain, to compare the outcomes.
Third, we use the knowledge stored in state-of-the-art medical ontologies to construct a comprehensive ontology of alternative medical terms, and we develop a method for simplifying medical text by extracting and replacing medical terms with layman alternatives.  To rank the alternatives, we define a scoring function that takes into account both the frequency of the replacement term and how well it fits into the sentence. Our experiments, using crowd-sourcing, show that our method is capable of simplifying complex medical text while retaining both its grammatically and meaning. 

We show that our method surpasses the state-of-the-art systems in medical text simplification, improving on grammaticality and meaning preservation of the simplified sentences. These aspects are particularly important in the context of medical text simplification, where factual correctness is paramount.

\makeatletter
\renewcommand{\@biblabel}[1]{\hfill #1.}
\makeatother

\bibliography{bibliography}

\begin{thebibliography}{10}

\bibitem{amrc}
Academy of~Medical Royal~Colleges.
\newblock Please, write to me. writing outpatient clinic letters to patients.
\newblock 2018.

\bibitem{Stajner2018}
Sanja {\v{S}}tajner and Sergiu Nisioi.
\newblock A detailed evaluation of neural sequence-to-sequence models for
  in-domain and cross-domain text simplification.
\newblock Miyazaki, Japan, May 2018. ELRA.

\bibitem{keselman2012classification}
Alla Keselman and Catherine~Arnott Smith.
\newblock A classification of errors in lay comprehension of medical documents.
\newblock {\em Journal of biomedical informatics}, 45(6):1151--1163, 2012.

\bibitem{van2019evaluating}
Laurens Van~den Bercken, Robert-Jan Sips, and Christoph Lofi.
\newblock Evaluating neural text simplification in the medical domain.
\newblock In {\em The World Wide Web Conference}, pages 3286--3292. ACM, 2019.

\bibitem{shardlow2019neural}
Matthew Shardlow and Raheel Nawaz.
\newblock Neural text simplification of clinical letters with a domain specific
  phrase table.
\newblock In {\em Proceedings of the 57th Annual Meeting of the ACL}, pages
  380--389, 2019.

\bibitem{Xu2015}
Wei Xu, Chris Callison-Burch, and Courtney Napoles.
\newblock Problems in current text simplification research: New data can help.
\newblock {\em Transactions of ACL}, 3:283--297, 2015.

\bibitem{Kauchak2013}
David Kauchak.
\newblock Improving text simplification language modeling using unsimplified
  text data.
\newblock In {\em Proceedings of the 51st Annual Meeting of the ACL (Volume 1:
  Long Papers)}, pages 1537--1546. ACL, August 2013.

\bibitem{10.1007/978-3-642-12320-7_5}
Lucia Specia.
\newblock Translating from complex to simplified sentences.
\newblock In {\em Computational Processing of the Portuguese Language}.
  Springer Berlin Heidelberg, 2010.

\bibitem{moses}
Philipp Koehn, Hieu Hoang, and Alexandra et~al. Birch.
\newblock Moses: Open source toolkit for statistical machine translation.
\newblock 06 2007.

\bibitem{coster_data}
William Coster and David Kauchak.
\newblock Simple english wikipedia: a new text simplification task.
\newblock In {\em Proceedings of the 49th Annual Meeting of the ACL: Human
  Language Technologies}, pages 665--669, 2011.

\bibitem{coster_lm}
William Coster and David Kauchak.
\newblock Learning to simplify sentences using wikipedia.
\newblock In {\em Proceedings of the workshop on monolingual text-to-text
  generation}, pages 1--9, 2011.

\bibitem{wubben}
Sander Wubben, Emiel Krahmer, and Antal Van~den Bosch.
\newblock Sentence simplification by monolingual machine translation.
\newblock volume~1, pages 1015--1024., 01 2012.

\bibitem{vstajner2015deeper}
Sanja {\v{S}}tajner, Hannah B{\'e}chara, and Horacio Saggion.
\newblock A deeper exploration of the standard pb-smt approach to text
  simplification and its evaluation.
\newblock In {\em ACL)}, 2015.

\bibitem{wang2016text}
Tong Wang, Ping Chen, John Rochford, and Jipeng Qiang.
\newblock Text simplification using neural machine translation.
\newblock In {\em Thirtieth AAAI Conference on Artificial Intelligence}, 2016.

\bibitem{nisioi-etal-2017-exploring}
Sergiu Nisioi, Sanja {\v{S}}tajner, Simone~Paolo Ponzetto, and Liviu~P. Dinu.
\newblock Exploring neural text simplification models.
\newblock Vancouver, Canada, July 2017. ACL.

\bibitem{opennmt}
Guillaume Klein, Yoon Kim, Yuntian Deng, Jean Senellart, and Alexander~M. Rush.
\newblock Open{NMT}: Open-source toolkit for neural machine translation.
\newblock In {\em Proc. ACL}, 2017.

\bibitem{sulem2018simple}
Elior Sulem, Omri Abend, and Ari Rappoport.
\newblock Simple and effective text simplification using semantic and neural
  methods.
\newblock {\em arXiv preprint arXiv:1810.05104}, 2018.

\bibitem{shardlow2014survey}
Matthew Shardlow.
\newblock A survey of automated text simplification.
\newblock {\em International Journal of Advanced Computer Science and
  Applications}, 4(1):58--70, 2014.

\bibitem{nelson2017unified}
Stuart~J Nelson, Tammy Powell, Suresh Srinivasan, and Betsy~L Humphreys.
\newblock Unified medical language
  system{\textregistered}(umls{\textregistered}) project.
\newblock In {\em Encyclopedia of library and information sciences}, pages
  4672--4679. CRC Press, 2017.

\bibitem{kohler2018expansion}
Sebastian Kohler, Leigh Carmody, Nicole Vasilevsky, Julius O~B Jacobsen, Danis,
  et~al.
\newblock Expansion of the human phenotype ontology (hpo) knowledge base and
  resources.
\newblock {\em Nucleic acids research}, 47(D1), 2018.

\bibitem{abrahamsson2014medical}
Emil Abrahamsson, Timothy Forni, Maria Skeppstedt, and Maria Kvist.
\newblock Medical text simplification using synonym replacement: Adapting
  assessment of word difficulty to a compounding language.
\newblock In {\em Workshop on Predicting and Improving Text Readability for
  Target Reader Populations}, 2014.

\bibitem{Uzuner2011}
Ozlem Uzuner, Brett South, Shuying Shen, and Scott DuVall.
\newblock 2010 i2b2/va challenge on concepts, assertions, and relations in
  clinical text.
\newblock {\em Journal of the American Medical Informatics Association :
  JAMIA}, 18:552--6, 06 2011.

\bibitem{Johnson2016}
Alistair~EW Johnson, Tom~J Pollard, Lu~Shen, H~Lehman Li-wei, Mengling Feng,
  and et~al. Ghassemi.
\newblock {MIMIC-III, a freely accessible critical care database}.
\newblock {\em Scientific Data}, 3(1):160035, 2016.

\bibitem{zheng2015entity}
Jin~G Zheng, Daniel Howsmon, Boliang Zhang, Juergen Hahn, Deborah McGuinness,
  James Hendler, and Heng Ji.
\newblock Entity linking for biomedical literature.
\newblock {\em BMC medical informatics and decision making}, 15(1):S4, 2015.

\bibitem{zeng2006exploring}
Qing~T Zeng and Tony Tse.
\newblock Exploring and developing consumer health vocabularies.
\newblock {\em Journal of the American Medical Informatics Association},
  13(1):24--29, 2006.

\bibitem{PMID:29632381}
Nicole~A Vasilevsky, Erin~D Foster, and et~all. Engelstad.
\newblock Plain-language medical vocabulary for precision diagnosis.
\newblock {\em Nature genetics}, 50(4):474—476, April 2018.

\bibitem{patwary2010experiments}
Md~Mostofa~Ali Patwary, Jean Blair, and Fredrik Manne.
\newblock Experiments on union-find algorithms for the disjoint-set data
  structure.
\newblock In {\em International Symposium on Experimental Algorithms}, pages
  411--423. Springer, 2010.

\bibitem{xu2016optimizing}
Wei Xu, Courtney Napoles, Ellie Pavlick, Quanze Chen, and Chris Callison-Burch.
\newblock Optimizing statistical machine translation for text simplification.
\newblock {\em Transactions of the ACL}, 4:401--415, 2016.

\bibitem{paetzold-specia-2016-semeval}
Gustavo Paetzold and Lucia Specia.
\newblock {S}em{E}val 2016 task 11: Complex word identification.
\newblock In {\em Proceedings of the 10th International Workshop on Semantic
  Evaluation}, pages 560--569, San Diego, California, June 2016. ACL.

\bibitem{chen1999empirical}
Stanley~F Chen and Joshua Goodman.
\newblock An empirical study of smoothing techniques for language modeling.
\newblock {\em Computer Speech \& Language}, 13(4):359--394, 1999.

\bibitem{gulordava2018colorless}
Kristina Gulordava, Piotr Bojanowski, Edouard Grave, Tal Linzen, and Marco
  Baroni.
\newblock Colorless green recurrent networks dream hierarchically.
\newblock {\em ACL}, 2018.

\bibitem{radford2019language}
Alec Radford, Jeffrey Wu, Rewon Child, David Luan, Dario Amodei, and Ilya
  Sutskever.
\newblock Language models are unsupervised multitask learners.
\newblock {\em OpenAI Blog}, 1(8), 2019.

\bibitem{radford2018improving}
Alec Radford, Karthik Narasimhan, Tim Salimans, and Ilya Sutskever.
\newblock Improving language understanding by generative pre-training.
\newblock 2018.

\bibitem{Papineni:2002:BMA:1073083.1073135}
Kishore Papineni, Salim Roukos, Todd Ward, and Wei-Jing Zhu.
\newblock Bleu: A method for automatic evaluation of machine translation.
\newblock In {\em Proceedings of the 40th Annual Meeting on ACL}, ACL '02,
  pages 311--318, 2002.

\bibitem{zhu2010monolingual}
Zhemin Zhu, Delphine Bernhard, and Iryna Gurevych.
\newblock A monolingual tree-based translation model for sentence
  simplification.
\newblock ACL, 2010.

\bibitem{doi:10.1177/1745691610393980}
Michael Buhrmester, Tracy Kwang, and Samuel~D. Gosling.
\newblock Amazon's mechanical turk: A new source of inexpensive, yet
  high-quality, data?
\newblock {\em Perspectives on Psychological Science}, 6(1):3--5, 2011.
\newblock PMID: 26162106.

\bibitem{celikyilmaz2020evaluation}
Asli Celikyilmaz, Elizabeth Clark, and Jianfeng Gao.
\newblock Evaluation of text generation: A survey.
\newblock 2020.

\bibitem{amidei2019use}
Jacopo Amidei, Paul Piwek, and Alistair Willis.
\newblock The use of rating and likert scales in natural language generation
  human evaluation tasks: A review and some recommendations.
\newblock 2019.

\bibitem{sulem-etal-2018-bleu}
Elior Sulem, Omri Abend, and Ari Rappoport.
\newblock {BLEU} is not suitable for the evaluation of text simplification.
\newblock In {\em Proceedings of the 2018 Conference on EMNLP}, pages 738--744,
  Brussels, Belgium, October-November 2018.

\end{thebibliography}
\bibliographystyle{unsrt}

\end{document}